\documentclass[twoside,leqno,twocolumn]{article}

\usepackage[letterpaper]{geometry}
\usepackage{ltexpprt}

\usepackage{amsmath,amssymb,amsfonts}
\usepackage{caption}
\usepackage{graphicx}
\usepackage{booktabs}
\usepackage{algorithm}
\usepackage{algorithmic}
\usepackage{amsfonts} 
\usepackage{amsmath, bm}
\usepackage{multirow}
\usepackage{bbm}
\usepackage{color}
\usepackage{subcaption}
\usepackage{adjustbox}
\usepackage{cite}
\usepackage{times}
\usepackage{url}

\begin{document}

\title{Fairness-aware Agnostic Federated  Learning}
\author{Wei Du \and Depeng Xu
\and Xintao Wu \thanks{University of Arkansas, \{wd005, depengxu, xintaowu\}@uark.edu} \and Hanghang Tong \thanks{University of Illinois, htong@illinois.edu}}

\date{}

\maketitle

\begin{abstract}
Federated learning is an emerging framework that builds centralized machine learning models with training data distributed across multiple devices. Most of the previous works about federated learning focus on the privacy protection and communication cost reduction. However, how to achieve fairness in federated learning is under-explored and challenging especially when testing data distribution is different from training distribution or even unknown. Introducing simple fairness constraints on the centralized model cannot achieve model fairness on unknown testing data. In this paper, we develop a fairness-aware agnostic federated learning framework (AgnosticFair) to deal with the challenge of unknown testing distribution. We use kernel reweighing functions to assign a reweighing value on each training sample in both loss function and fairness constraint. Therefore, the centralized model built from AgnosticFair can achieve high accuracy and fairness guarantee on unknown testing data.  Moreover, the built model can be directly applied to local sites as it guarantees fairness on local data distributions. To our best knowledge, this is the first work to achieve fairness in federated learning. Experimental results on two real datasets demonstrate the effectiveness in terms of both utility and fairness under data shift scenarios.
\end{abstract}

\section{Introduction}
In traditional machine learning, the training data is usually stored in a central server. The central server needs to first collect the data from different sources and combines these data together to facilitate the learning. The rapid development of the various machine learning or deep learning models benefits significantly from the large-scale training data. However, the above training approach also raises data privacy concerns  because the raw data needs to be uploaded to the server, which can lead to the possible sensitive information leakage.

To alleviate the above issues caused by centralized learning, federated learning is proposed as an attractive framework to handle complex data \cite{mcmahan2016communication,smith2017federated,yang2019federated}, as it enables a new implementation of distributed deep learning over a large number of clients. Compared to the traditional centralized learning which collects all local data samples and  builds the model at a central server, federated learning trains local models on local data samples and local clients exchange parameters to generate a global model. Although tremendous research has been done on the federated learning, how to achieve fairness in federated learning is under-explored. Fairness receives increasing attentions in machine learning. Previous research  demonstrates that many machine learning models are often biased or unfair against some protected groups especially when they were trained on biased data. How to achieve fairness in federated learning is  more urgent because the training data used in federated learning is often geo-distributed among various groups.  

One challenge of achieving fairness in federated learning is due to the statistical challenge of the unknown testing data distribution. In this paper, we consider a supervised task with features $\textbf{X}$ and labels $Y$ and assume $P(y | \textbf{x})$ is common across all clients. Our goal is to train a single global model that learns $P(y | \textbf{x})$. The single global model can then be shared to clients or provided to new clients with no training data. However, in federated learning, the distributions of training data at different clients are often different, i.e., $P_i (\mathbf{x},y) \neq P_j (\mathbf{x},y)$. Data shift clearly exists between the training distribution of local data used to build the model and the unknown testing distribution. This data shift causes significant challenges for developing fair federated learning algorithms because the learned model may have poor performance on testing data and the learned model with fairness constraint on the training data cannot guarantee the fairness on the testing data. 

It is beneficial to build a learning model that is robust against the possible unknown testing distribution in terms of both utility and fairness. The authors in \cite{mohri2019agnostic} propose agnostic federated learning to deal with the unknown testing data distribution. They model the testing distribution as a mixture of the client data distributions and the mixture weight of one client is deviated from the proportion of its local data in the whole training data. They define the agnostic empirical loss with mixture weights and present a fast stochastic optimization algorithm. However, in their formulation, the model is optimized for the performance of the single worst client and does not take the fairness into consideration.  

In this work, we propose fairness-aware agnostic federated learning (AgnosticFair) to deal with the challenge when the testing data distribution is unknown. We formulate AgnosticFair as a two-player adversarial minimax game between the learner and the adversary. The adversary aims to generate any possible unknown testing data distribution to maximize the classifier loss. We assign an individual reweighing value on each training sample and incorporate reweighing value in both agnostic loss function and agnostic fairness constraint. 
As a result, the global model learned in the minimax game achieves both high accuracy and fairness guarantee on unknown testing data. Moreover, each client can simply deploy the global model on its local site as the learned global model guarantees fairness on any local data. We conduct extensive experiments on two public datasets and compare our approach with several baselines. Evaluation results demonstrate the effectiveness of our proposed AgnosticFair in terms of accuracy and fairness under data shift scenarios.   
The main contributions of this work are summarized as follows:
\begin{itemize}
    \item To the best of our knowledge, our research is the first one that formulates the problem of fairness-aware federated learning under the data distribution shift among the clients.
    \item We propose to use  kernel function parametrization in  loss function and fairness constraints so that they are both agnostic to the data distribution shift.
    \item We develop an efficient approach to optimize the agnostic loss function under the agnostic fairness constraints between the server and clients. During the optimization process, only parameters and coefficients are needed to exchange between clients and the server without disclosing any raw data.
    \item We conduct extensive experiments to demonstrate that our approach can achieve fair prediction under distribution shift while maintaining high accuracy. 
\end{itemize}

The rest of the paper is organized as follows. Section \ref{sec:relatedwork} reviews the related work regarding the federated learning and fairness machine learning. Section \ref{sec:formulation} presents the problem formulation of fairness agnostic federated learning and techniques to solve the problem. Section \ref{sec:experiment} shows the experimental results of our proposed framework. Section \ref{sec:conclusion} concludes the paper and presents the future work.

\section{Related Work}
\label{sec:relatedwork}
\subsection{Federated Learning}
There exist extensive research works on federated learning aiming to solve various challenges. The pioneer work of federated learning is proposed by \cite{mcmahan2016communication} where it considers that data is distributed among different mobile devices and proposes an algorithm called FedAvg to enable the collaborative learning among different clients. In this pioneer work, it points out several challenges mainly including the limited bandwidth of federated learning clients, privacy leakage and non-IID (independent and identically distributed) data among different clients. Lots of subsequent works investigate on how to solve the above issues.

The work \cite{mcmahan2016communication} takes the first shoot to reduce massive communication cost by applying FedAvg to reduce the communication rounds during the model training. The authors in \cite{lin2017deep} propose to sparsify the exchanged parameters by selecting important ones so that the total number of parameters per round is reduced significantly. In \cite{alistarh2017qsgd}, the authors design a gradient quantization approach using less bits to represent the full 32 bits float gradient, which can reduce the communication cost in both uploading and downloading stages. In addition, there are also lots of works toward reducing communication cost \cite{suresh2017distributed,sun2019communication,zheng2019communication, chen2018adacomp, mills2019communication}. The privacy protection in federated learning also receives increasing attentions in the past few years. For example, the work \cite{bonawitz2017practical} encrypts the exchanged parameters of local clients before uploading to the central server. The authors in \cite{agarwal2018cpsgd} adopt the concept of differential privacy and add differentially private noise on the exchanged parameters to protect the client privacy. Applying the encryption and differential privacy to prevent privacy leakage are also reported in \cite{shokri2015privacy,truex2019hybrid,papernot2016semi,wang2019beyond}.

The non-IID data widely exists in federated learning due to the massive distributed clients and our work falls into this category.
In \cite{smith2017federated} the authors study the federated learning in a multi-task setting where each client collects individual data with its own statistical pattern. The proposed model learns individual pattern for each client while simultaneously borrow shared information from other clients. The research \cite{li2019convergence} considers the non-IID distribution among different clients and provides theoretical convergence analysis. The authors in \cite{mansour2020three} point out the distribution shift in the training data of federated learning and  proposes three approaches to adapt the federated learning model to enable the personalization of each local client. In \cite{zhao2018federated} the authors propose a federated learning framework to train the model from non-IID data, where a small subset of data is globally shared among all the clients. 
Most recently, the authors in \cite{mohri2019agnostic} propose agnostic federated learning. However, the proposed framework solves the problem by optimizing the worst case for a single client. There are extensive research works on addressing distribution shift between training and testing data in the centralized learning. In \cite{grunwald2004game} the authors formulate a two-player minimax game by optimizing the worst case of expected loss and introduce using  kernel reweighing functions to address unknown distribution shift. Our research is inspired by this work. Reweighing methods are proposed to deal with the known distribution shift when the training and testing data are available. For example,  density ratio estimation is used to reweigh training data to represent known testing data \cite{shimodaira2000improving},  a sample reweighing approach is designed to correct the distribution shift \cite{huang2007correcting}, and reweighing values are used in loss optimization \cite{wen2014robust,pan2010domain}. 

\subsection{Fairness in Machine Learning}

Fairness-aware learning receives lots of attentions in the past few years. 
Typically fairness is defined to protect some specific attributes and there are two major directions to achieve fairness. The first type of work is either pre-processing the training data to remove sensitive information about the protected attribute or post-processing the classifier  to achieve fair prediction. For example, the work in \cite{feldman2015certifying} studies the disparate impact of the prediction on protected class and proposes to construct a paired version of the original training dataset to remove the disparate impacts on protected class. In \cite{zhang2017achieving} the authors develop a causal framework to discover and remove the discriminatory effects on the training data. The research in  \cite{xu2018fairgan} applies generative adversarial neural networks to generate fair data from the original training data and uses the generated data to train the model.

The second type of work is to incorporate the fairness constraint into the classification model during the optimization process \cite{zafar2017fairness,donini2018empirical,zafar2015fairness,cotter2019optimization,hashimoto2018fairness,woodworth2017learning,baharlouei2019r} and our work also falls into this direction. For instance, the work \cite{woodworth2017learning} adds a non-discrimination constraint on the training samples during the optimization of the classifier. The authors in \cite{cotter2019optimization} study the optimization of non-convex and non-differentiable constraints induced by the fairness. However, most of the previous works on fairness learning are in the centralized setting and do not consider distribution shift. 
Most recently, the authors in \cite{li2019fair} propose an algorithm to achieve fair accuracy distribution across devices in federated networks. It minimizes a reweighted loss that assigns high weights for devices with higher losses, where the central concept is to achieve fair accuracy for different devices. The authors in \cite{mohri2019agnostic} mention concept of fairness in federated learning but do not propose any algorithms to achieve it. The authors in \cite{hashimoto2018fairness} consider the fairness in an online learning setting across different groups. Similar to \cite{mohri2019agnostic}, it also aims to minimize the worst case of a single group. In short, our work firstly formulates the problem of fairness-aware federated learning under the data distribution shift and develops an effective algorithm that  uses kernel function parametrization in both loss function and fairness constraints to achieve fairness guarantee and good accuracy under unknown testing data. 

\begin{table}[!t]
\centering \caption{Notations} \label{table:notation}
\scalebox{0.8}{
\begin{tabular}[t]{|c|l|} \hline
Symbol & Definition \\ \hline
$u_1, u_2, \dots, u_p$ & $p$ clients in federated learning   \\ \hline
$\mathcal{D}_k$ & local dataset of $u_k$  \\ \hline
$t_i^k =(\mathbf{x}_i^k, y_i^k)$ & the $i$-th tuple of $u_k$  \\ \hline
$S$ & sensitive attribute  \\ \hline
$\mathbf{w}$ & the parameter vector of federated learning model \\ \hline
$L(\textbf{w})$ & loss function of federated learning model \\ \hline
$L(\textbf{w}, \bm{\alpha})$ & loss function of agnostic federated learning model \\ \hline
$l(f_k(\textbf{x}_i^k; \textbf{w}), y_i^k)$ & loss value of $t_i^k$ in $u_k$\\ \hline
$\bm{\alpha}$ & coefficients of reweighing function\\ \hline
$\theta_{\bm{\alpha}}(\textbf{x})$& reweighing function \\ \hline
$K(\textbf{x})$ & Gaussian kernel function\\ \hline
$C_{\mathcal{D}}(\bm{\alpha};\textbf{x}; \textbf{w})$ & agnostic decision boundary fairness constraint\\ \hline
$\textbf{w}_k^t$ & parameters of $u_k$ at $t$th step\\ \hline
$\bar{\textbf{w}}^t$ & average parameters of $p$ users at $t$th step\\ \hline
$\bm{\alpha}^t$ & values of $\bm{\alpha}$ at $t$th step\\ \hline
$L, \theta, C$ & loss, equality, inequality constraint in Equation \ref{eq:federated_learning_minimax_fairness}\\ \hline
$\phi_L^t, \phi_C^t$ & coefficient vector of $\textbf{w}$ of $L, C$ at $t$th step \\ \hline
$\phi_{L,k}^t, \phi_{C,k}^t$ & coefficient vector of $\textbf{w}$ of $L, C$ of $u_k$ at $t$th step \\ \hline
$\psi_{L}^t,\psi_{\theta}^t, \psi_{C}^t$ & coefficient vector of $\bm{\alpha}$ of $L, \theta, C$ at $t$th step  \\ \hline
$\psi_{L,k}^t,\psi_{\theta,k}^t, \psi_{C,k}^t$ & coefficient vector of $\bm{\alpha}$ of $L, \theta, C$ of $u_k$ at $t$th step  \\ \hline
\end{tabular}}
\end{table}

Different from all previous works, we first study the fairness-aware classification in federated learning. The most relevant works are \cite{hashimoto2018fairness} and \cite{mohri2019agnostic}, however, both \cite{hashimoto2018fairness} and \cite{mohri2019agnostic} consider the minimization of a single client or group, which limits the number of clients or groups to smaller scale. In contrast, we develop a general framework by using a rewighing function on individual sample to solve the data distribution shift issue among different clients. From this perspective, \cite{hashimoto2018fairness} and \cite{mohri2019agnostic} are a special case of our framework. We also construct a robust fairness constraint that can achieve both high accuracy and fairness on unknown testing data.

\section{Fair Agnostic Federated Learning}
\label{sec:formulation}
\subsection{Problem Formulation}
We first define the following notations used throughout the paper. Suppose there exist $p$ local clients $u_1, u_2, \dots, u_p$ in the federated learning setting and each client is associated with a dataset $\mathcal{D}_k = \{\textbf{X}, Y\}, k \in [1, p]$. 
The $k$th client contains $n_k$ samples 
and each sample is denoted as $t_i^k:\{\textbf{x}_i^k, y_i^k\}, i \in [1, n_k]$. The total number of data samples is defined as $n = \sum_{k = 1}^p n_k$. Let $\textbf{X} \in \mathcal{X}$ be the input space and $Y \in \mathcal{Y}$ be the output space. We consider the binary classification that $\mathcal{Y} = \{0, 1\}$.  The global model $f$ predicts the label as $\hat{y} = f(\textbf{x})$. 
The goal of the federated learning is to collaboratively train a machine learning model $f$ by these $p$ clients. 

The  standard  federated  learning  framework aims to minimize the empirical risk $L(\textbf{w})$ over all records and learns the parameter vector $\textbf{w} \in \mathcal{W}$ as the following:
\begin{equation}\label{eq:standard_federated_learning_loss}
\begin{aligned}
&\min_{\textbf{w}\in \mathcal{W}} L(\textbf{w}) =  \dfrac{1}{n}\sum_{k = 1}^p 
\sum_{i = 1}^{n_k} l(f_k(\textbf{x}_{i}^k; \textbf{w}), y_i^k)
\end{aligned}
\end{equation}
where $f_k$ is the classifier of $u_k$, $l(f_k(\textbf{x}_i^k; \textbf{w}), y_i^k)$ is the loss of sample $t_i^k$ in $u_k$. In general, federated learning includes the following steps. 
\begin{itemize}
    \item Step 1: The server specifies the learning task, e.g., linear regression or deep neural network, for these $p$ clients and sends out initial parameters $\textbf{w}^0$.
    \item Step 2: At $t$th step, $u_k$ receives averaged parameters $\bar{\textbf{w}}^t$ from the server as the new round of initialization parameters and then finds out the optimal ${\textbf{w}}_k^{t+1}$ using its local data $\mathcal{D}_k$:
    \begin{equation}\label{eq:local_learning_loss}
        \textbf{w}_k^{t+1} = \text{arg} \min_{\textbf{w}\in \mathcal{W}}  \dfrac{1}{n_k} \sum_{i = 1}^{n_k} l(f_k(\textbf{x}_{i}^k; \textbf{w}), y_i^k).
    \end{equation}
    \item Step 3: The server receives parameters $\textbf{w}_k^{t+1}$ from each client $u_k$ and sends averaged parameter $\bar{\textbf{w}}^{t+1} = \dfrac{1}{p}\sum_{k = 1}^{p}\textbf{w}_k^{t+1}$ back to all clients. 
\end{itemize}
It should be noted that step 2 and step 3 are repeated until reaching the preset convergence. We present the pseudo code of the standard federated learning framework in Algorithm \ref{alg:FLFramework} in order to compare with our fairness-aware agnostic federated learning in Algorithm \ref{alg:AFLFramework}.

\begin{algorithm}[t]   
	\caption{Federated Learning Framework}   
	\label{alg:FLFramework}   
	\begin{algorithmic}[1] 
		\REQUIRE ~~\\ 
		 $\mathcal{D}_k$ from client $u_k $, $k=1,\cdots,p$;\\
		Training steps $T$; \\
		Initial parameter vector $\textbf{w}^0$; \\
		\ENSURE ~~\\Global model parameter vector $\textbf{w}$; 
		\STATE Initialize parameters $\textbf{w}^0$ for all clients; 
		\STATE $t = 0$;
		\STATE \textbf{While $t \leq T$ do}
		\STATE \hspace{0.2cm}\textbf{Client Side}:
		\STATE \hspace{0.2cm}\textbf{for} $k = 1 : p$ \textbf{do}
		\STATE \hspace{0.5cm}Client $k$ receives averaged $\bar{\textbf{w}}^t$ and computes $\textbf{w}_k^{t+1}$  \\ \hspace{0.5cm}using Equation \ref{eq:local_learning_loss};
        \STATE \hspace{0.5cm}Client $k$ uploads $\textbf{w}_k^{t+1}$ to server;
		\STATE \hspace{0.2cm}\textbf{Server Side}:
		\STATE \hspace{0.5cm}Server receives $\textbf{w}_k^{t+1}$ ($1 \leq k \leq p$) from all clients;
		\STATE \hspace{0.5cm}Server computes $\bar{\textbf{w}}^{t+1} = \dfrac{1}{p}\sum_{k = 1}^{p}\textbf{w}_k^{t+1}$ and sends \\ \hspace{0.5cm}back to each client;
		\STATE \hspace{0.3cm}$t = t + 1$;
		\STATE \textbf{return} $\bar{\textbf{w}}^{T}$
	\end{algorithmic}
\end{algorithm}

In the fair learning, without loss of generality, we assume $S$ is one sensitive attribute in $\textbf{X}$ with $S = 0$ representing the minority group and $S = 1$  the majority group. Following the standard federated learning framework, we write the objective function subject to the fairness constraint: 
\begin{equation}\label{eq:federated_learning_loss}
\begin{aligned}
&\min_{\textbf{w}\in \mathcal{W}} L(\textbf{w}) =  \dfrac{1}{n}\sum_{k = 1}^p 
\sum_{i = 1}^{n_k} l(f_k(\textbf{x}_{i}^k; \textbf{w}), y_i^k)\\
&\text{subject to}\quad g(\textbf{x}; \textbf{w}) \leq \epsilon,
\end{aligned}
\end{equation}
where $g(\textbf{x}; \textbf{w}) \leq \epsilon$ is the fairness constraint and related to the sensitive attribute $S$. The weight of each sample in Equation \ref{eq:federated_learning_loss} from these $p$ clients is uniform. The underlying assumption of the standard federated learning framework is that the testing data distribution is the same as the distribution of the training data (union of data samples from $p$ clients). However, this assumption is rather restrictive and will lead to the following possible drawbacks. First, the performance of the trained model will be degraded if the distribution of the training data and that of the testing data do not coincide. Second, the fairness achieved on the training data does not guarantee the fairness on the testing data.

\subsection{Agnostic Loss Function}
It is usually considered that the distribution shift exists between the training data $P_{tr}(\textbf{X})$ and the testing data $P_{te}(\textbf{X})$, whereas the conditional distribution $P(Y|\textbf{X})$ indicating the prediction remains the same. To correct the distribution shift between $P_{tr}(\textbf{X})$ and $P_{te}(\textbf{X})$, a widely used approach is to reweigh the training samples in the learning process so that the learned model $f$ can reflect the testing data distribution. 
We write the objective function in Equation \ref{eq:federated_learning_loss} with the reweighed training samples as the following:
\begin{equation}\label{eq:federated_learning_reweigh_loss}
\min_{\textbf{w} \in \mathcal{W}} L(\textbf{w}) =  \dfrac{1}{n}\sum_{k = 1}^p 
\sum_{i = 1}^{n_k} \theta(\textbf{x}_i^k)l(f_k(\textbf{x}_{i}^k; \textbf{w}), y_i^k),
\end{equation}
where $\theta(\textbf{x})$ is the reweighing function to correct the distribution shift from $P_{tr}(\textbf{X})$ to $P_{te}(\textbf{X})$. 
There exist several methods to estimate the reweighing value $\theta(\textbf{x})$ if the (unlabelled) testing data is given \cite{quionero2009dataset}. For example, applying density ratio estimation can compute the reweighing value  $\theta(\textbf{x}) = P_{te}(\textbf{x})/P_{tr}(\textbf{x})$ for each example, then the reweighing value can represent the possible testing distribution in real scenarios.

However, we cannot properly estimate the reweighing values if we do not have available testing data. Therefore, it is necessary to extend the above framework by building a classifier which is favorable to any unknown testing distribution. We define the agnostic loss over any unknown testing data as the following:
\begin{equation}\label{eq:federated_learning_reweigh_minimax_loss}
\begin{aligned}
&\min_{\textbf{w} \in \mathcal{W}} \hspace{0.1cm} \max_{\theta \in \Theta} \hspace{0.1cm} L(\textbf{w}, \theta) = \dfrac{1}{n}\sum_{k = 1}^p \sum_{i = 1}^{n_k} \theta(\textbf{x}_i^k)l(f_k(\textbf{x}_{i}^k; \textbf{w}), y_i^k).
\end{aligned}
\end{equation}
$\Theta$ represents the set of unknown testing data distribution produced by the adversary. The formulation can be considered as  a two-player adversarial game such that the adversary in Equation \ref{eq:federated_learning_reweigh_minimax_loss} tries to select a reweighing function $\theta \in \Theta$ to maximize the loss of the objective, whereas the learner tries to find parameters  $\textbf{w} \in \mathcal{W}$ to minimize the worst case loss over the unknown testing data distribution produced by the adversary.

The proposed framework by Equation \ref{eq:federated_learning_reweigh_minimax_loss} enjoys several advantages. First, under the  independent and identically distributed (IID) data settings, the minimization of the robust reweighed loss is equivalent and dual to the empirical risk minimization (objective function in Equation \ref{eq:federated_learning_loss}) \cite{grunwald2004game}. It indicates that the optimization of Equation \ref{eq:federated_learning_reweigh_minimax_loss} will not cause performance degradation when no distribution shift exists. Second, the optimal $\textbf{w} \in \mathcal{W}$ is minimized for the worst case loss and the performance of the global model is robust with any unknown testing data.

\subsection{Kernel Function Parametrization}
The reweighing function $\theta \in \Theta$ can be chosen based on the prior knowledge or the application scenario. A reweighing function on individual data sample across clients usually corrects the distribution shift more accurately. 
We rewrite the agnostic loss in Equation \ref{eq:federated_learning_reweigh_minimax_loss} as:
\begin{equation}\label{eq:federated_learning_reweigh_minimax_loss_gaussian_kernel}
\begin{aligned}
&\min_{\textbf{w} \in \mathcal{W}} \hspace{0.05cm} \max_{\bm{\alpha} \in \textbf{R}^{+}}   \hspace{0.1cm} L(\textbf{w}, \bm{\alpha}) = \dfrac{1}{n}\sum_{k = 1}^p \sum_{i = 1}^{n_k} \theta_{\bm{\alpha}}(\textbf{x}_i^k)l(f_k(\textbf{x}_{i}^k; \textbf{w}), y_i^k)\\
&\text{subject to}  \hspace{0.1cm}\dfrac{1}{n}\sum_{k = 1}^p \sum_{i = 1}^{n_k} \theta_{\bm{\alpha}}(\textbf{x}_i^k) = 1.
\end{aligned}
\end{equation}
$\theta_{\bm{\alpha}}(\textbf{x})$ is the reweighing function that is linearly parametrized as the following:
\begin{equation}\label{eq:reweighing_alpha}
    \theta_{\bm{\alpha}}(\textbf{x}) = \sum_{m=1}^M\alpha_m K_m(\textbf{x}), 0 \leq \alpha_m \leq B
\end{equation}
where $K_m(\textbf{x})$ is a basis function, $M$ is the number of basis functions, and $\bm{\alpha}$ contains the mixing coefficients $\alpha_1, \alpha_2, \cdots, \alpha_M$. 
The sum-to-one constraint of the reweighing function $\theta_{\bm{\alpha}}(\textbf{x})$ ensures that it can properly model the unknown distribution shift from the training data to the testing data. The coefficient $\alpha_m$ is non-negative and bounded by $B \in \textbf{R}^{+}$, which constrains the value of $\theta_{\bm{\alpha}}(\textbf{x})$ and controls the capacity of the adversary.
The linearly parametrized reweighing function $\theta_{\bm{\alpha}}(\textbf{x})$ has two advantages. First, the linear form allows us to choose multiple basis functions to capture many different possible uncertainties of the unknown testing data. Second, the optimization with linear form can be more easily solved by linear programming or convex programming tool.

In fact, there are many options to choose the form of basis functions. In this paper, we choose the Gaussian kernel
\begin{equation}\label{eq:Gaussian_kernel}
    K_m(\textbf{x}_i^k) = \text{exp}(-\parallel \textbf{b}_m - \textbf{x}_i^k\parallel^2/2\sigma^2)
\end{equation}
with the basis $\textbf{b}_m$ and the kernel width $\sigma$. The basis $\textbf{b}_m$ can be chosen based on some prior knowledge, for example, we can set $\textbf{b}_m$ as some possible centers of the testing data according to the hypothesis. $\sigma$ is the width of the kernel.
As the smaller variation of $\parallel \textbf{b}_m - \textbf{x}\parallel^2$ will cause larger value change of the kernel function and smaller $\sigma$ indicates more possible testing distributions that the adversary can generate \cite{wen2014robust}.  
Or the basis $\textbf{b}_m$ could be an indicator function $\mathbbm{1}_{[.]}$ which represents groups from different clients, ages, or domains. The value generated by each kernel function can be seen as a conditional probability $P(\textbf{x}|m)$ of observing $\textbf{x}$ given the class $m$ in a mixture model. The mixing coefficients $\bm{\alpha} \in \mathcal{A}$ are usually bounded in the non-negative Euclidean space. 

The agnostic federated learning framework proposed by \cite{mohri2019agnostic} is a special case of our framework. As being said that the basis function could be an indicator function $\mathbbm{1}_{[\cdot]}$ representing a group. In \cite{mohri2019agnostic}, it models the testing data distribution as an unknown mixture of $p$ clients where each group is assigned with a uniform weight.
The unknown testing data distribution in \cite{mohri2019agnostic} can be constructed under our framework  as the following:
\begin{equation}
    \theta_{\bm{\alpha}}(\textbf{x}_i^k) = \lambda_k \dfrac{n_k}{n}, 1 \leq i \leq n_k, 1 \leq k \leq p
\end{equation}
where $\dfrac{n_k}{n}$ is the uniform weight of each data $\textbf{x}_i^k$ before reweighing.
More specifically, for each client $u_k$, the agnostic federated learning \cite{mohri2019agnostic} assigns the same value $\lambda_k$ to reweigh each data in $\mathcal{D}_k$. However, assigning reweighing value at the client level is insufficient to model the unknown distribution shift due to the following two reasons. First, the data from the same client also has diversity and needs different reweighing values. Second, different clients can have similar data and these similar data should be assigned with similar reweighing values. In our framework, we assign the reweighing value at the individual data level, which is more capable of modeling the unknown testing data distribution.

\subsection{Agnostic Fairness Constraint}
The fairness constraint $g(\textbf{x};\textbf{w})$ in standard federated learning 
(Equation \ref{eq:federated_learning_loss}) assigns uniform weight for each sample. When it comes to the unknown testing data, the fairness achieved by Equation \ref{eq:federated_learning_loss} may not guarantee the fairness on unknown testing data. As being said, the adversary tries to produce a set of possible unknown testing distributions. It encourages us to construct the agnostic fairness constraint based on the unknown testing distributions by the adversary. Then, the optimization of the objective function is subject to the fairness constraint based on the unknown testing distribution. 

There exist several notions for fairness $g(\textbf{x};\textbf{w})$ and demographic parity is the most widely used notion in fairness machine learning \cite{hardt2016equality}. It requires the prediction result by the model to be independent of the sensitive attribute $S$. Demographic parity is usually quantified by the risk difference, which measures the difference between positive predictions on the sensitive and the non-sensitive groups. The risk difference of a classifier $f$ is expressed as:
\begin{equation}
    RD(f) = |P(\hat{Y} = 1 | S = 1) - P(\hat{Y} = 1 | S = 0)|,
\end{equation}
where $\hat{Y}$ is the predicted value of $f$.
For each client, we define $\mathcal{D}_{ij}^{k} = \{\textbf{x}|\hat{Y} = i, S = j\}$ where $i,j \in [0, 1]$. For notation convenience, we define $\mathcal{D}_{\cdot j}^k = \{\textbf{x}|S = j\}$ where $j \in [0, 1]$ and $\cdot$ represents $\{0, 1\}$.
Then we can write the expression for $ RD(f)$ with uniform weight on training data
as the following:
\begin{equation}\label{eq:risk_difference}
    |\dfrac{\sum_{k = 1}^p\sum \mathbbm{1}_{\textbf{x}_i^k \in \mathcal{D}_{11}^k}}{\sum_{k = 1}^p\sum\mathbbm{1}_{\textbf{x}_i^k \in \mathcal{D}_{\cdot 1}^k}} - \dfrac{\sum_{k = 1}^p\sum \mathbbm{1}_{\textbf{x}_i^k \in \mathcal{D}_{10}^k}}{\sum_{k = 1}^p\sum\mathbbm{1}_{\textbf{x}_i^k \in \mathcal{D}_{\cdot 0}^k}}| \leq \epsilon,
\end{equation}
where $\mathbbm{1}_{[.]}$ is an indicator function and $\epsilon \in [0, 1]$ is a threshold for the fairness constraint. However, Equation \ref{eq:risk_difference} is constructed based on the training data and cannot preserve fairness on the unknown testing data. Hence, we use the same reweighing function to construct the agnostic fairness constraint as the following:
\begin{equation}\label{eq:reweighted_risk_difference}
    |\dfrac{\sum_{k = 1}^p\sum_{\textbf{x}_i^k \in \mathcal{D}_{11}^k}\theta_{\bm{\alpha}}(\textbf{x}_i^k)}{\sum_{k = 1}^p\sum_{\textbf{x}_i^k \in \mathcal{D}_{\cdot 1}^k}\theta_{\bm{\alpha}}(\textbf{x}_i^k)} - \dfrac{\sum_{k = 1}^p\sum_{\textbf{x}_i^k \in \mathcal{D}_{10}^k}\theta_{\bm{\alpha}}(\textbf{x}_i^k)}{\sum_{k = 1}^p\sum_{\textbf{x}_i^k \in \mathcal{D}_{\cdot 0}^k}\theta_{\bm{\alpha}}(\textbf{x}_i^k)}| \leq \epsilon.
\end{equation}
Then our proposed fairness-aware agnostic federated learning (AgnosticFair) is the combination of Equation \ref{eq:federated_learning_reweigh_minimax_loss_gaussian_kernel} and Equation \ref{eq:reweighted_risk_difference}. The fairness constraint in Equation \ref{eq:reweighted_risk_difference} is constructed based on the unknown testing distribution, so when it comes to the unknown testing data, the trained classifier can still preserve the fairness. Another benefit is that even though the distributions of the local clients and the server side do not coincide, the classifier can still guarantee the fairness on each local client due to the agnostic fairness constraint.

The optimal solution of Equation \ref{eq:federated_learning_reweigh_minimax_loss_gaussian_kernel} under the fairness constraint by Equation \ref{eq:reweighted_risk_difference} is computationally intractable to obtain because the fairness constraint contains the indicator function. An alternative fairness constraint is defined as the covariance between the sensitive attribute and the signed distance from the non-sensitive attribute vector to the decision boundary. It has been proved that the decision boundary fairness is a concept of risk difference \cite{wu2019convexity}. We write this alternative definition $C_{\mathcal{D}}(\textbf{x}; \textbf{w})$ as:
\begin{equation}\label{eq:fairness_convariance}
    C_{\mathcal{D}}(\textbf{x}; \textbf{w}) = \dfrac{1}{n}\sum_{k=1}^p\sum_{i = 1}^{n_k}(s_{\textbf{x}_i^k} - \bar{s})d_{\textbf{w}(\textbf{x}_i^k)},
\end{equation}
where $s_{\textbf{x}_i^k}$ is the value of the sensitive attribute of the sample $\textbf{x}_i^k$, $d_{\textbf{w}(\textbf{x}_i^k)}$ is the distance to the decision boundary of the classifier $f$, $\bar{s}$ is the mean value of the sensitive attribute over $\mathcal{D}$ that is $ \dfrac{\sum_{k=1}^p\sum_{i=1}^{n_k} s_{\textbf{x}_i^k}}{n}$. To achieve fair classification, it is required that $|C_{\mathcal{D}}(\textbf{x}; \textbf{w})| \leq \tau$ where $\tau \in \textbf{R}^+$. 
Incorporating the reweighing values into the fairness constraint gives:
\begin{equation}\label{eq:reweigh_fairness_convariance}
    C_{\mathcal{D}}(\bm{\alpha};\textbf{x}; \textbf{w}) = \dfrac{1}{n}\sum_{k=1}^p\sum_{i = 1}^{n_k}(s_{\textbf{x}_i^k} - \bar{s} )\theta_{\bm{\alpha}}(\textbf{x}_i^k)d_{\textbf{w}(\textbf{x}_i^k)}.
\end{equation}

\subsection{Solving Fairness-aware Agnostic Federated Fairness Learning}

Now we are ready to formulate our  agnostic federated learning under the decision boundary fairness constraint as:
\begin{equation}\label{eq:federated_learning_minimax_fairness}
\begin{aligned}
\min_{\textbf{w} \in \mathcal{W}} \hspace{0.1cm} \max_{\bm{\alpha} \in \textbf{R}^{+}} \quad & L(\textbf{w}, \bm{\alpha}) = \dfrac{1}{n}\sum_{k = 1}^p\sum_{i = 1}^{n_k} \theta_{\bm{\alpha}}(\textbf{x}_i^k) l(f_k(\textbf{x}_i^k; \textbf{w}), y_i^k)\\
\text{subject to} \quad & \dfrac{1}{n}\sum_{k = 1}^p \sum_{i = 1}^{n_k} \theta_{\bm{\alpha}}(\textbf{x}_i^k) = 1, 0 \leq \alpha_m \leq B
\\ & |\dfrac{1}{n}\sum_{k=1}^p\sum_{i = 1}^{n_k}(s_{\textbf{x}_i^k} - \bar{s} )\theta_{\bm{\alpha}}(\textbf{x}_i^k)d_{\textbf{w}(\textbf{x}_i^k)}| \leq \tau.
\end{aligned}
\end{equation}

The optimization of Equation \ref{eq:federated_learning_minimax_fairness} includes two sets of parameters, $\bm{\alpha}$ and $\textbf{w}$. The minimax expression encourages us to alternatively optimize $\bm{\alpha}$ and $\textbf{w}$
in an iterative way. The client and server will collaboratively optimize $\textbf{w}$ and $\bm{\alpha}$ to solve the minimax problem. The general pipeline is that the client optimizes $\textbf{w}$ with fixed $\bm{\alpha}$, while the server optimizes $\bm{\alpha}$ with fixed $\textbf{w}$. One challenge is how both the server and the clients conduct optimization iteratively through sharing parameters or intermediate results (rather than raw data), as required in federated learning.

The objective loss $L(\textbf{w}, \bm{\alpha})$ (abbreviated as $L$) shown in Equation \ref{eq:federated_learning_minimax_fairness} can be written as a function of $\bm{w}$ with corresponding coefficients $\phi_L$ when $\bm{\alpha}$ is fixed. More importantly, the second summation over samples in client $u_k$, $\sum_{i = 1}^{n_k} \theta_{\bm{\alpha}}(\textbf{x}_i^k) l(f_k(\textbf{x}_i^k; \textbf{w}), y_i^k)$, can be similarly expressed as a function $\bm{w}$ with corresponding coefficients  $\phi_{L,k}$. We can easily see $\phi_L = \sum_{k= 1}^{p}\phi_{L,k}$ and hence each client can simply send $\phi_{L,k}$ (rather than any raw data) to the server. Similarly, when $\bm{w}$ is fixed, $L(\textbf{w}, \bm{\alpha})$ 
can be written as a function of $\bm{\alpha}$ with corresponding coefficients $\psi_L$,  the second summation over samples in client $u_k$ has coefficients  $\psi_{L,k}$, and  $\psi_L = \sum_{k= 1}^{p}\psi_{L,k}$. 

\begin{figure*}
\centering
\includegraphics[scale = 0.65]{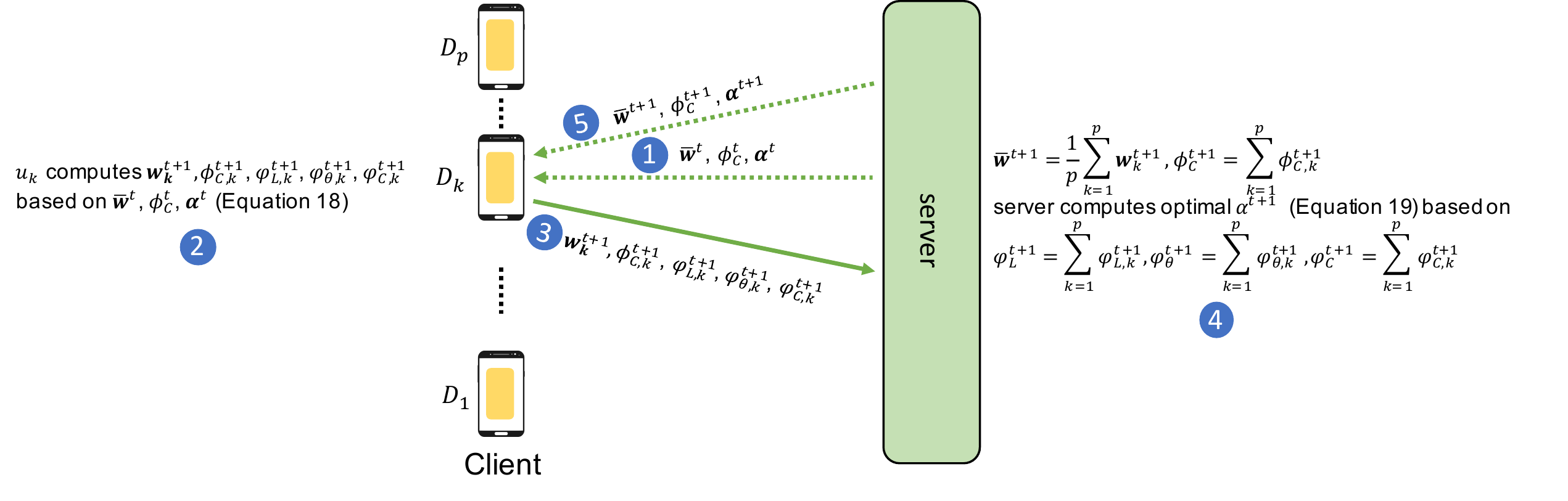}
\caption{The interaction between the client and server in federated learning. The client side optimizes its local $\textbf{w}$ and server side optimizes the $\bm{\alpha}$.}
\label{fig:federated_learning}
\end{figure*}

Similarly, the equality constraint for the reweighing functions (abbreviated as $\theta$) and the inequality constraint for the decision boundary fairness (abbreviated as $C$) in Equation \ref{eq:federated_learning_minimax_fairness} can be expressed as functions with corresponding coefficients. 
Throughout this paper, we use $\phi$ to denote the coefficients vector of $\textbf{w}$ (with fixed $\alpha$), $\psi$ to denote the coefficients vector of $\alpha$ (with fixed $\textbf{w}$). We use the subscript $L$, $\theta$ and $C$ denote the loss function, equality constraint, and inequality constraint, and further add the subscript $k$ for the coefficients from client $u_k$. Moreover, we use the superscript $t$ to express the coefficients at step $t$ during the optimization. For example, $\psi_{C,k}^t$ denotes the coefficient vector of $\alpha$ (with fixed $\textbf{w}$) in the 
inequality constraint formula for client $u_k$ at step $t$. We note that the equality constraint only involves variable $\bm{\alpha}$ and does not have coefficient vector $\phi_\theta$. 
We put all of the notations in Table \ref{table:notation} and show their relationships in the following equation. 
\begin{equation}\label{eq:coefficient}
\begin{aligned}
    \quad&\phi_L^t = \sum_{k= 1}^{p}\phi_{L,k}^t, \phi_{C}^t = \sum_{k= 1}^{p}\phi_{C,k}^t \\
    &\psi_L^t = \sum_{k= 1}^{p}\psi_{L,k}^t, \psi_{\theta}^t = \sum_{k= 1}^{p}\psi_{\theta,k}^t, \psi_{C}^t = \sum_{k= 1}^{p}\psi_{C,k}^t. \\
\end{aligned}
\end{equation}

\begin{algorithm}[t]   
	\caption{AgnosticFair: Fairness-aware Agnostic Federated Learning}   
	\label{alg:AFLFramework}   
	\begin{algorithmic}[1] 
		\REQUIRE ~~\\ 
	    $\mathcal{D}_k$ from client $u_k $, $k=1,\cdots,p$;\\
		Training steps $T$; \\
		Initial parameters $\textbf{w}^0$ and $\bm{\alpha}^0$; \\
		\ENSURE ~~\\Global model parameter vector $\textbf{w}$; 
		\STATE Initialize parameters $\textbf{w}^0$ and $\bm{\alpha}^0$ for all clients; 
		\STATE $t = 0$;
		\STATE \textbf{While $t \leq T$ do}
		\STATE \hspace{0.2cm}\textbf{Client Side}:
		\STATE \hspace{0.2cm}\textbf{for} $k = 1 : p$ \textbf{do}
		\STATE \hspace{0.5cm}Client $k$ receives averaged ${\bar{\textbf{w}}}^t$, $\bm{\alpha}^{t}$ and $\phi_C^t$; \\
		\hspace{0.5cm}Client $k$ computes optimal $\textbf{w}_k^{t+1}$ using Equation \ref{eq:agnostic_objective_fairness_penalty_term} \\\hspace{0.5cm} and uploads to server;
        \STATE \hspace{0.5cm}Client $k$ computes $\phi_{C,k}^t, \psi_{L,k}^{t+1},\psi_{\theta,k}^{t+1},\psi_{C,k}^{t+1}$ and uploads \\\hspace{0.5cm}to server;
		\STATE \hspace{0.2cm}\textbf{Server Side}:
		\STATE \hspace{0.5cm}Server aggregates
		$\psi_L^{t+1} = \sum_{k= 1}^{p}\psi_{L,k}^{t+1}$, \\\hspace{0.5cm}$ \psi_{\theta}^{t+1} = \sum_{k= 1}^{p}\psi_{\theta,k}^{t+1}, \psi_{C}^{t+1} = \sum_{k= 1}^{p}\psi_{C,k}^{t+1}$;
        \STATE \hspace{0.5cm}Server computes optimal $\bm{\alpha}^{t+1}$ using Equation \ref{eq:federated_learning_optimize_alpha};
		\STATE \hspace{0.5cm}Server aggregates $\phi_C^{t+1} = \sum_{
k= 1}^{p}\phi_{C,k}^{t+1}$ and averages \\\hspace{0.5cm} $\bar{\textbf{w}}^{t+1} = \dfrac{1}{p}\sum_{k = 1}^{p}\textbf{w}_k^{t+1}$;
		\STATE \hspace{0.5cm}Server sends back $\bar{\textbf{w}}^{t+1}$, $\bm{\alpha}^{t+1}$ and $\phi_C^{t+1}$;\\
		\STATE \hspace{0.3cm}$t = t + 1$;
		\STATE \textbf{return} $\bar{\textbf{w}}^{T}$
	\end{algorithmic}
\end{algorithm}

In the following, we present details about the optimization process between the server and client. We show those key parameters and coefficients exchanged between the client and the server as well as their calculations in  Figure \ref{fig:federated_learning}.

\vspace{0.2cm}
\noindent \textbf{Client side}:
In standard federated learning, each client computes $\textbf{w}$ based on its local data and exchanges the updated $\textbf{w}$ with other clients via the server. Here we also follow this standard approach that each client computes the optimal values of $\textbf{w}$ locally using the fixed $\bm{\alpha}$ received from the server. We can decompose the part of Equation \ref{eq:federated_learning_minimax_fairness} related to $\textbf{w}$ as the following:
\begin{equation}\label{eq:federated_learning_optimize_w}
\begin{aligned}
\min_{\textbf{w} \in \mathcal{W}} \hspace{0.1cm} \quad & L(\textbf{w}) = \dfrac{1}{n}\sum_{k = 1}^p\sum_{i = 1}^{n_k} \theta_{\bm{\alpha}}(\textbf{x}_i^k) l(f_k(\textbf{x}_i^k; \textbf{w}), y_i^k)\\
\text{subject to} \quad & |\dfrac{1}{n}\sum_{k=1}^p\sum_{i = 1}^{n_k}(s_{\textbf{x}_i^k} - \bar{s} )\theta_{\bm{\alpha}}(\textbf{x}_i^k)d_{\textbf{w}(\textbf{x}_i^k)}| \leq \tau.
\end{aligned}
\end{equation}
It can be seen that given the fixed $\bm{\alpha}$, the optimization of $\textbf{w}$ is only subject to the inequality constraint. The optimization of Equation \ref{eq:federated_learning_optimize_w} depends on the choice of loss function and the learning model. For example, the loss function of linear regression is convex and the inequality constraint of $\textbf{w}$ is also linear so the convex programming tool can be used to solve it. However, the loss function over $\textbf{w}$ of many other machine learning models (e.g., deep learning models) is not convex. Hence, it is challenging to optimize the non-convex function subject to the constraint. We observe that the inequality constraint in Equation \ref{eq:federated_learning_optimize_w} is used to guarantee the fairness of the updated $\textbf{w}$ during the optimization. Instead of optimizing non-convex loss function subject to the constraints, we can transform the inequality constraint to a penalty term on the loss function. 

We choose the square term for fairness inequality constraint as a penalty and rewrite the loss function of $\textbf{w}$ for client $k$ as the following:
\begin{equation}\label{eq:agnostic_objective_fairness_penalty_term}
\begin{aligned}
\min_{\textbf{w} \in \mathcal{W}} L(\textbf{w}) = \quad &\dfrac{1}{n_k}\sum_{i = 1}^{n_k}[\theta(\textbf{x}_i^k) l(f_k(\textbf{x}_i^k; \textbf{w}), y_i^k)] \\&+ \lambda(\dfrac{1}{n}\sum_{k=1}^p\sum_{i = 1}^{n_k}(s_{\textbf{x}_i^k} - \bar{s} )\theta_{\bm{\alpha}}(\textbf{x}_i^k)d_{\textbf{w}(\textbf{x}_i^k)} - \tau)^2
\end{aligned}
\end{equation}
where $\lambda$ is a hyperparameter controlling the trade-off between the classification accuracy and the fairness. Equation \ref{eq:agnostic_objective_fairness_penalty_term} includes two terms. The first term is the loss of each client based on its own data while the second term is the global fairness constraint.

Suppose client $u_k$ receives the average parameters $\bar{\textbf{w}}^t$ and $\bm{\alpha}^t$ from the server at the $t$th step, it can compute $\phi_{L,k}^t$ using $\bm{\alpha}^t$ and local data $\mathcal{D}_k$. For the second term computation, it needs to receive $\phi_{C}^t = \sum_{k= 1}^{p}\phi_{C,k}^t$ from the server, where each client can compute $\phi_{C,k}^t$ independently using local data $\mathcal{D}_k$. In this process, the raw data of each client is not exposed, which fulfills the requirement of federated learning. 
Given $\bar{\textbf{w}}^t$, $\phi_{L,k}^t$ and $\phi_{C}^t$, client
$u_k$ obtains the complete form of Equation \ref{eq:agnostic_objective_fairness_penalty_term} and can compute optimal ${\textbf{w}_k^{t+1}}$ based on $\mathcal{D}_k$. Based on ${\textbf{w}_k^{t+1}}$ and fixed $\bm{\alpha}^t$, it can compute $\psi_{L,k}^{t+1},\psi_{\theta,k}^{t+1},\psi_{C,k}^{t+1}$, and $\phi_{C,k}^{t+1}$ and upload them to the server.

\vspace{0.2cm}
\noindent \textbf{Server side}: The optimization of $\bm{\alpha}$ is subject to both equality and inequality constraints, which is expressed as:
\begin{equation}\label{eq:federated_learning_optimize_alpha}
\begin{aligned}
\max_{\bm{\alpha} \in \textbf{R}^{+}} \quad & L(\bm{\alpha}) = \dfrac{1}{n}\sum_{k = 1}^p\sum_{i = 1}^{n_k} \theta_{\bm{\alpha}}(\textbf{x}_i^k) l(f_k(\textbf{x}_i^k; \textbf{w}), y_i^k)\\
\text{subject to} \quad & \dfrac{1}{n}\sum_{k = 1}^p \sum_{i = 1}^{n_k} \theta_{\bm{\alpha}}(\textbf{x}_i^k) = 1, 0 \leq \alpha_m \leq B
\\ & |\dfrac{1}{n}\sum_{k=1}^p\sum_{i = 1}^{n_k}(s_{\textbf{x}_i^k} - \bar{s} )\theta_{\bm{\alpha}}(\textbf{x}_i^k)d_{\textbf{w}(\textbf{x}_i^k)}| \leq \tau.
\end{aligned}
\end{equation}
Given fixed $\textbf{w}$, $\theta_{\bm{\alpha}}$ is a linear function subject to linear equality and inequality constraints.
The server aggregates coefficient vector  $\psi_{L}^{t+1},\psi_{\theta}^{t+1},\psi_{C}^{t+1}$ of $\bm{\alpha}$ (Equation \ref{eq:coefficient}) to obtain the complete form of Equation \ref{eq:federated_learning_optimize_alpha}.
The server then uses linear programming tool to obtain the optimal values $\bm{\alpha}^{t+1}$ and sends back to each client. In addition, the server also  averages parameters $\bar{\textbf{w}}^{t+1} = \dfrac{1}{p}\sum_{k = 1}^{p}\textbf{w}_k^{t+1}$, aggregates  $\phi_{C}^{t+1} = \sum_{k=1}^{p}\phi_{C,k}^{t+1}$, and sends them back to each client for next round iteration.

We also present the pseudo code of our proposed fairness-aware agnostic federated learning (AgnosticFair) in Algorithm \ref{alg:AFLFramework}. It can be seen that each client optimizes $\textbf{w}$ at the local side and the server optimizes $\bm{\alpha}$. The final classifier with fair prediction is achieved through the iterative optimization process.

\subsection{Variants of AgnosticFair}

There exist several variants of our fairness-aware  agnostic federated learning which can be used as baselines. We consider the following two variations in this paper and will show their experimental results in Section \ref{sec:experiment}.

The first variation is termed as \text{AgnosticFair-a} that optimizes agnostic loss (Equation \ref{eq:federated_learning_reweigh_minimax_loss_gaussian_kernel}) without any fairness constraint.  \text{AgnosticFair-a} only deals with the data distribution shift regarding the accuracy and has no fairness guarantee on the model. The goal of considering \text{AgnosticFair-a} is to test its accuracy performance with comparison to \cite{mohri2019agnostic} using our proposed reweighing function. 

The second variation is termed as \text{AgnosticFair-b} that considers the agnostic loss and uniform weighted fairness constraint, which is expressed as the following:
\begin{equation}\label{eq:federated_learning_AgnosticFair-b}
\begin{aligned}
\min_{\textbf{w} \in \mathcal{W}} \hspace{0.1cm} \max_{\bm{\alpha} \in \textbf{R}^{+}} \quad & L(\textbf{w}, \bm{\alpha}) = \dfrac{1}{n}\sum_{k = 1}^p\sum_{i = 1}^{n_k} \theta_{\bm{\alpha}}(\textbf{x}_i^k) l(f_k(\textbf{x}_i^k; \textbf{w}), y_i^k)\\
\text{subject to} \quad & \dfrac{1}{n}\sum_{k = 1}^p \sum_{i = 1}^{n_k} \theta_{\bm{\alpha}}(\textbf{x}_i^k) = 1, 0 \leq \alpha_m \leq B
\\ &|\dfrac{\sum_{k = 1}^p\sum \mathbbm{1}_{\textbf{x}_i^k \in \mathcal{D}_{11}^k}}{\sum_{k = 1}^p\sum\mathbbm{1}_{\textbf{x}_i^k \in \mathcal{D}_{\cdot 1}^k}} - \dfrac{\sum_{k = 1}^p\sum \mathbbm{1}_{\textbf{x}_i^k \in \mathcal{D}_{10}^k}}{\sum_{k = 1}^p\sum\mathbbm{1}_{\textbf{x}_i^k \in \mathcal{D}_{\cdot 0}^k}}| \leq \epsilon.
\end{aligned}
\end{equation}
It is expected that \text{AgnosticFair-b} can guarantee the fairness under the IID setting but fails to achieve fairness under unknown distribution shift. For comparison, our AgnosticFair (Equation \ref{eq:federated_learning_minimax_fairness}) can guarantee fairness while maintain high accuracy performance under the unknown distribution shift. We will show their comparisons in Section \ref{sec:experiment}. 

\section{Experiments}
\label{sec:experiment}
\subsection{Experimental Setup}
\textbf{Datasets.} We evaluate our proposed approach AgnosticFair on two datasets, Adult dataset \cite{Dua:2019} and Dutch dataset \cite{vzliobaite2011handling}. Adult dataset collects the personal information from different people including age, education level, race, gender, and so forth. The prediction task is to determine whether the income of a person is over 50K or not. Dutch dataset collects personal information of the inhabitants in Netherlands and the task is also to classify the individual into high income or low income. For both datasets, we set “gender” as the sensitive attribute. For non-sensitive attributes, we apply one-hot encoding to convert the categorical attributes into vectors and normalize numerical attributes to the range within $[0, 1]$. After preprocessing, Adult dataset consists of 45222 data samples and each data sample has 40 features, whereas Dutch dataset consists of 60420 data samples and each data sample has 35 features. 

To create the distribution shift scenarios from the training set to the testing set, we artificially split each dataset as the following. For Adult, the training set $\mathcal{D}^{tr}$ contains 80\% data of people working in private company and 20\% data of people working in other organizations, and the testing set $\mathcal{D}^{te}$ contains the rest of the data. Hence, $\mathcal{D}^{tr}$ of Adult is dominated by data of people working in private company, while $\mathcal{D}^{te}$ of Adult is dominated by data of people working in other organizations. We consider 2 local clients in our experiment $u_1$ and $u_2$. $u_1$ only contains data of people working in private company from $\mathcal{D}^{tr}$ while $u_2$ only contains data of people working in other groups. 
Similarly for Dutch, the training set $\mathcal{D}^{tr}$ contains 80\% data of people who are married with children and 40\% data of people from other groups, and $\mathcal{D}^{te}$ contains the rest of the data. We also consider 2 local clients, $u_1$ contains data of people who are married with children while $u_2$ contains of people in other groups.

\begin{table*}
  \centering
  \caption{Model performance under data distribution shift (Adult and Dutch) Acc: accuracy}
  \scalebox{1.0}{
  \begin{tabular}{|c|c|c|c|c|c|c|}
  \hline
	\multirow{2}*{\textbf{Methods}} & \multicolumn{3}{|c|}{\textbf{Adult Dataset}} & \multicolumn{3}{|c|}{\textbf{Dutch Dataset} }\\
	\cline{2-7}
     & {Training Acc} & {Testing Acc} & {Testing $RD$} & {Training Acc} & {Testing Acc} & {Testing $RD$} \\
    \hline
    \hline
     \text{FL}   &  0.7500 &  0.7998 &  0.1477 & 0.8133 &  0.7951 & 0.1945\\
     \hline
     \text{AgnosticFair} &  0.7413 &  0.7626 &  0.0196 & 0.7478 & 0.7205 & 0.0371\\
     \hline
     \text{AgnosticFair-a} &  0.7820 &  0.8294 &  0.1306 & 0.8259 & 0.8162 & 0.2154\\
     \hline
     \hline
     \text{FairFL} &  0.7537 &  0.7534 &  0.0852 & 0.6899 & 0.7011  & 0.0961\\
     \hline
     \text{\cite{mohri2019agnostic}} & 0.7761  & 0.7774  & 0.1150  & 0.8170 & 0.7738 & 0.1238 \\
     \hline
  \end{tabular}}
  \label{tab: accuracy_data_shift}
\end{table*}

\begin{table*}
  \centering
  \caption{Local fairness and global fairness under data distribution shift (Adult)}
  \scalebox{1.0}{
  \begin{tabular}{|c|c|c|c|c|}
  \hline
	\multicolumn{1}{|c|}{\textbf{Methods}}& {$u_1$ Testing $RD$} & {$u_2$ Testing $RD$} & {Global Testing $RD$} & {Global Testing Accuracy}\\
    \hline
     \text{AgnosticFair} & 0.0208 & 0.0177  & 0.0196 & 0.7626\\
     \hline
     \text{AgnosticFair-b} &  0.0450  & 0.0795 & 0.0673 & 0.7885\\
     \hline
  \end{tabular}}
  \label{tab: local_fairness_data_shift}
\end{table*}

\begin{table*}
  \centering
  \caption{Model performance of IID data(Adult and Dutch) Acc: accuracy}
  \scalebox{1.0}{
  \begin{tabular}{|c|c|c|c|c|c|c|}
  \hline
	\multirow{2}*{\textbf{Methods}} & \multicolumn{3}{|c|}{\textbf{Adult Dataset}} & \multicolumn{3}{|c|}{\textbf{Dutch Dataset} }\\
	\cline{2-7}
     & {Training Acc} & {Testing Acc} & {Testing $RD$} & {Training Acc} & {Testing Acc} & {Testing $RD$} \\
    \hline
    \hline
     FL   &  0.8129 &  0.8130 &  0.1490 & 0.8116 & 0.8096  & 0.1698 \\
     \hline
     AgnosticFair &  0.7938 &  0.7749 &  0.0299 & 0.7338 & 0.7322 & 0.0270 \\
     \hline
     AgnosticFair-a &  0.8083 &  0.8111 &  0.1515 & 0.8135 & 0.8089 & 0.1526\\
     \hline
     \hline
     FairFL &  0.7731 &  0.7723 &  0.0235 & 0.7564 & 0.7346  & 0.0325\\
     \hline
     \cite{mohri2019agnostic} & 0.7774  &  0.7785 & 0.1484  & 0.7925 & 0.7892 & 0.1547 \\
     \hline
  \end{tabular}}
  \label{tab: random_split_data}
\end{table*}

\noindent \textbf{Hyperparameters.} In our experiment, we use Gaussian kernel in Equation \ref{eq:Gaussian_kernel} as the reweighing function to construct the unknown testing data distribution. The upper bound $B$ for $\bm{\alpha}$ is set as 5 and $\sigma$ is chosen to be 1. In fact, the upper bound of $B$ is rarely reached in practical optimization so that it will not limit the power of the adversary too much. The basis of the Gaussian kernel is chosen from training data and the number of kernels is set as 200. The threshold $\tau$ in Equation \ref{eq:federated_learning_minimax_fairness} is set as a constant 0.05 and $\lambda$ in Equation \ref{eq:agnostic_objective_fairness_penalty_term} is set as 2.

\noindent \textbf{Baselines.} In our experiment, we use the logistic regression model to evaluate our proposed algorithm. We compare the performance of our proposed \text{AgnosticFair} with the following baselines: (a) standard federated learning (\text{FL}) without fairness constraint; (b) standard federated learning with fairness constraint (FairFL); (c) agnostic federated learning \cite{mohri2019agnostic} that assigns the reweighing value at the client level. To conduct meaningful comparison between our model with baselines, we introduce several variations: \text{AgnosticFair-a} that optimizes agnostic loss (Equation \ref{eq:federated_learning_reweigh_minimax_loss_gaussian_kernel}) without any fairness constraint; \text{AgnosticFair-b} that optimizes agnostic loss subject to  unweighted fairness constraint (Equation \ref{eq:federated_learning_AgnosticFair-b}).

\noindent \textbf{Metrics.} We evaluate our proposed framework and baselines based on utility and fairness. We use accuracy to measure the utility and risk difference ($RD$) to measure the fairness. A fair classifier usually has $RD(f) \leq 0.05$. We run all experiments 20 times and report the average results. 

\subsection{Comparison under Unknown Data Shift}

\begin{figure}
\centering
\includegraphics[scale = 0.4]{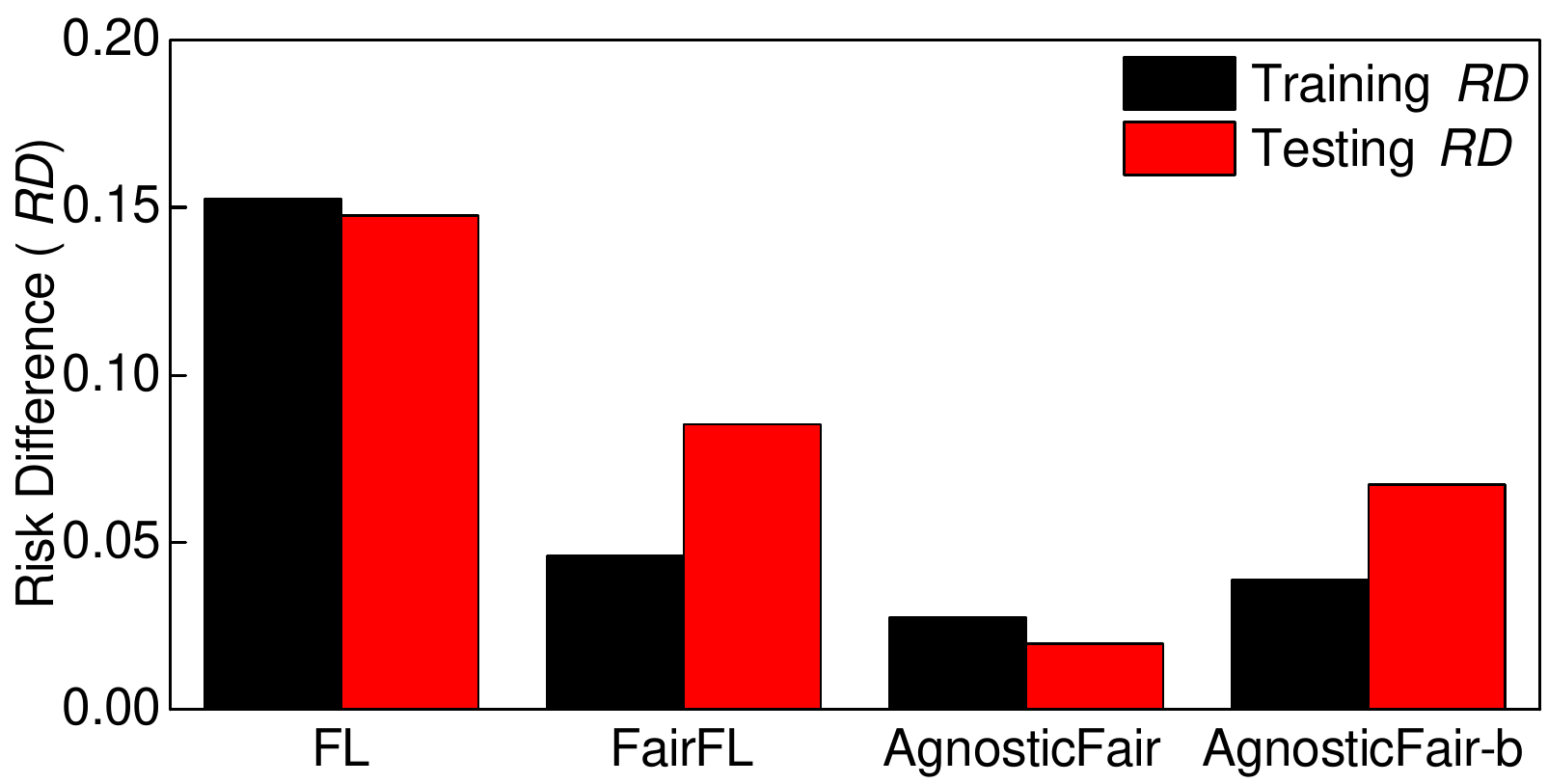}
\caption{Model Fairness under data distribution shift (Adult)}
\label{fig:angnostic_fair}
\end{figure}

\begin{figure}
\centering
\includegraphics[scale = 0.35]{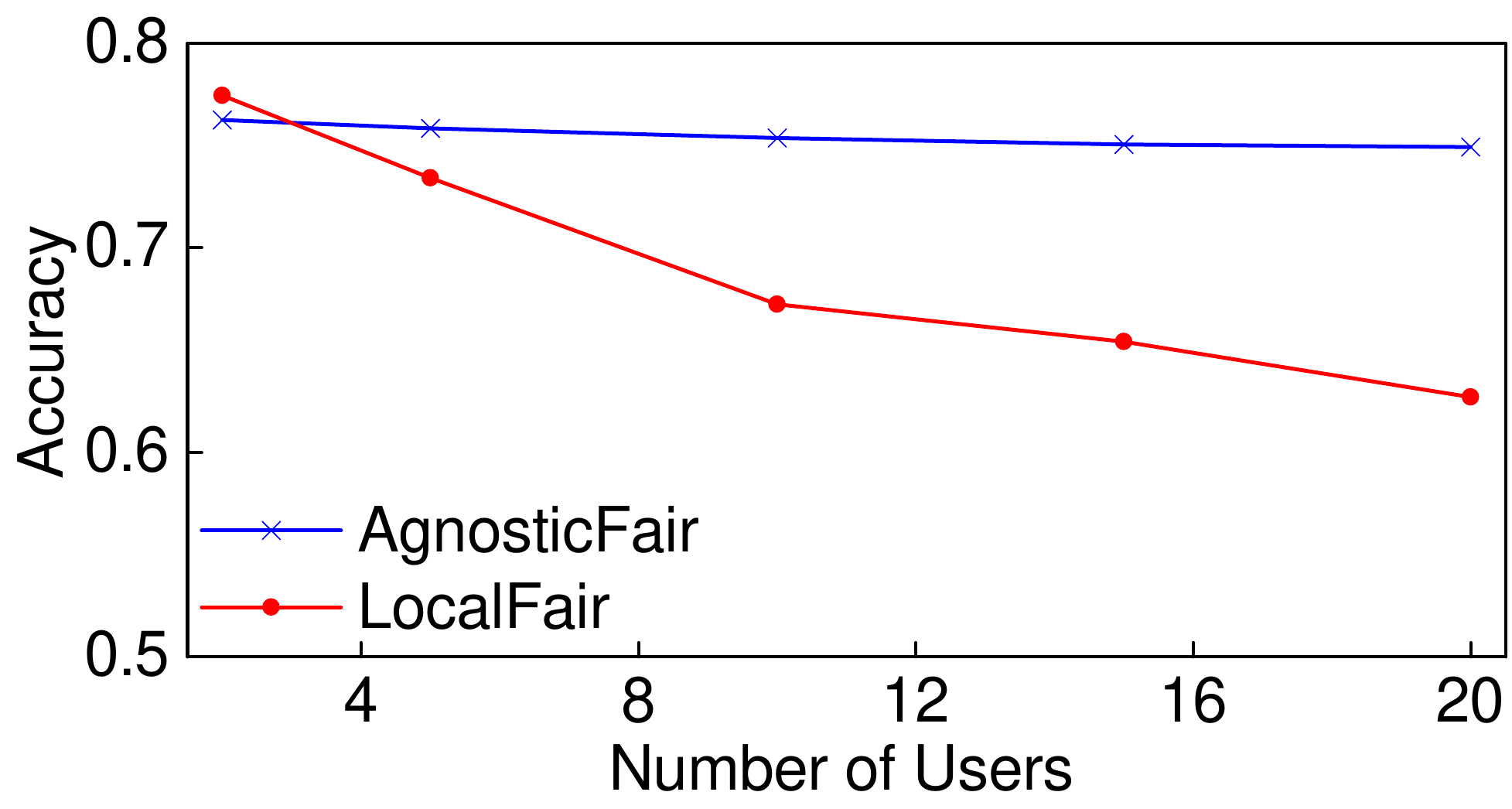}
\caption{The accuracy of the global model with different number of local clients (Adult).}\label{fig:local_fair_vs_angnostic_fair}
\end{figure}

In our framework, the reweighing function is designed to improve the performance of the classifier if the data distributions of the training and the testing set do not coincide. The use of the reweighing function in the fair constraint can also achieve fairness guarantee under unknown testing data.  

\noindent \textbf{Accuracy.} We report the experimental results in Table \ref{tab: accuracy_data_shift} that demonstrate the accuracy improvement by our framework.
We summarize several interesting points regarding accuracy as the following. 
The testing accuracy of \text{AgnosticFair-a} is 0.8264 on Adult and 0.8162 on Dutch, while the accuracy of \text{FL} is 0.7998 on Adult and 0.7951 on Dutch. More specifically, \text{AgnosticFair-a} outperforms \text{FL} by 0.0296 on Adult and by 0.0211 on Dutch. It demonstrates that the agnostic loss function in Equation \ref{eq:federated_learning_reweigh_minimax_loss} improves the performance of the model when it comes to the distribution change. 

Compared to \cite{mohri2019agnostic}, \text{AgnosticFair-a} enjoys higher accuracy. As we discussed before, assigning a reweighing value at the client level optimizes the client with the worst loss which will reduce its generalization ability on the unknown testing data. For \text{AgnosticFair-a}, we assign a reweighing value for each data sample across all clients and put more weights on difficult data samples with higher loss. In fact, the difficult data samples can come from any client and are taken into the consideration during the training process. Hence, the generalization ability in the testing stage of \text{AgnosticFair-a} is increased.

\noindent \textbf{Fairness.} 
In this experiment, we will show that our proposed AgnosticFair can achieve fairness guarantee under unknown  data distribution shift. In Table \ref{tab: accuracy_data_shift}, the two columns ``Testing RD'' show the risk difference values of our Agnostic-Fair and four baselines over the testing data of both Adult and Dutch. We also draw a plot  in Figure \ref{fig:angnostic_fair} to show achieving fairness on the training data by  baselines cannot guarantee fairness on testing data whereas our AgnosticFair can achieve the guarantee.
First, experimental results show that both \text{AgnosticFair} and \text{AgnosticFair-b} can achieve fairness on the training data, but only \text{AgnosticFair} can guarantee the fairness on the testing data. \text{AgnosticRegFair-b} uses the agnostic loss function, but its fairness constraint is unweighted. It can be concluded that using the agnostic loss function only cannot guarantee the fairness when it comes to the unknown testing data. 
Second, \text{FL} achieves high accuracy but cannot achieve the fairness. The $RD$ of the \text{FL} is 0.1477, whereas a fair learning model usually requires $RD$ to be less than $0.05$. The fairness constraint of \text{FairFL} does not consider data distribution shift. The results show that \text{FairFL} achieves fairness on the training data but fails on the testing data. 

\noindent \textbf{Federated Learning with Different Number of Clients.} Our proposed AgnosticFair can also achieve fairness for local clients when the distribution shift exists between the local clients and the global server. In fact, the distribution shift from the global training data to the local client data is a special case of the unknown testing data distribution. We use the same data split setting and report the result of local fairness and global fairness in Table \ref{tab: local_fairness_data_shift}. It can be seen that \text{AgnosticFair-b} cannot guarantee the fairness on the unknown testing data because it fails on $u_2$ ($RD = 0.0795$). Due to the agnostic fairness constraint of \text{AgnosticFair}, it can achieve fairness for both local clients. To demonstrate the stability of our proposed \text{AgnosticFair}, we show its accuracy under different number of clients in Figure \ref{fig:local_fair_vs_angnostic_fair}. We use the same data split setting and distribute the data evenly to each client without overlap. The accuracy of the global classifier is recorded when fairness is achieved on all clients. It can be seen that the performance of \text{AgnosticFair} is independent of the number of clients. For comparison, we also use another straightforward approach \text{LocalFair} that achieves fairness for each local client by adding a local fairness constraint based on its own data. It can guarantee the fairness for local client, however, the drawback is to add a local fairness constraint for each client, which will reduce the utility of the global model if more clients are included. Figure \ref{fig:local_fair_vs_angnostic_fair} also shows the accuracy curve of \text{LocalFair}, we can see that its performance degrades significantly with the increasing number of clients.

\subsection{Comparison under IID Setting}
In our last experiment, we also test the performance of our model under the IID data setting. In this experiment, we randomly split two datasets, Adult and Dutch. For each dataset, we use 80\% of the data as the training set and the rest 20\% as the testing set. The number of local clients is set to be 2 and the training data is evenly distributed to each local client. Table \ref{tab: random_split_data} shows the experimental results.

First, \text{AgnosticFair-a} achieves same level of performance with \text{FL} if no data distribution shift exists. For Adult, the testing accuracy of \text{FL} (\text{AgnosticFair-a}) is 0.8130 (0.8111). For Dutch, the testing accuracy of \text{FL} (\text{AgnosticFair-a}) is 0.8096 (0.8089). It is quite interesting because in \cite{grunwald2004game} the authors state: under (IID) data, the minimization of the robust reweighed loss (Equation \ref{eq:federated_learning_reweigh_loss} in our framework) is equivalent and dual to the empirical risk minimization (Equation \ref{eq:federated_learning_loss} in our framework). Hence, the accuracy of the \text{AgnosticFair-a} and \text{FL} also echoes the theoretical statement in \cite{grunwald2004game}.

Second, our proposed \text{AgnosticFair-a} also achieves higher accuracy than \cite{mohri2019agnostic} under the IID setting. More specially, the testing accuracy is 0.7785 (0.7892) for Adult (Dutch), which is still lower than that of \text{FL}. As aforementioned, \cite{mohri2019agnostic} assigns a different weight for each client. The optimization process as stated in their work will improve the worst loss of the individual client, whereas the global generalization on the testing data will be weakened. Finally, our \text{AgnosticFair} achieves fairness guarantees while preserve good accuracy. 

\section{Conclusions and Future Work}
\label{sec:conclusion}

In this paper, we have proposed a fairness-aware agnostic federated learning framework to deal with unknown testing data distributions. We apply  kernel reweighing functions to parametrize the loss function and fairness constraint. Hence our framework can achieve both good model accuracy and fairness on unknown testing data. We  conducted a series of experiments on two datasets and  experimental results  demonstrated three benefits of the trained centralized model by our fairness-aware agnostic federated learning. First, it can improve the prediction accuracy under the distribution shift from the training data to the testing data. Second, it can guarantee fairness on the unknown testing data. Third, it can guarantee the fairness of each local client. In our future work, we will extend our framework to cover other commonly used fairness notations. e.g., equalized odds and equalized opportunity \cite{hardt2016equality}, and incorporate surrogate functions in  agnostic fair constraints of our framework to  address the challenge of the indicator function used in fairness notations. We will also study kernel function parametrization with different basis functions.  Our proposed framework can also be adapted to the centralized fairness-aware learning where the training and testing data differ. Moreover, the proposed framework can also be applied in the fair transfer learning where  distribution shift usually exists between the source domain and target domain.

\section{Acknowledgement}
This work was supported in part by NSF  1920920, 1937010 and 1946391.

\end{document}